\documentclass{article}

\usepackage{amsmath}
\usepackage{array}
\usepackage{booktabs}
\usepackage{float}
\usepackage{graphicx}
\usepackage{multirow}
\usepackage{xcolor}
\usepackage{hyperref}

\usepackage[preprint]{corl_2026} 

\title{R2RDreamer: 3D-aware Data Augmentation for Spatially-generalized 2D Manipulation Policies}

\author{
Xiuwei Xu$^{1*}$,
  Haowen Sun$^{1*}$,
  Angyuan Ma$^{1*}$,
  Yiwei Zhang$^{1}$,
  Zhenyu Wu$^{2}$, \\
  \textbf{Xiaofeng Wang}$^{3}$,
  \textbf{Bingyao Yu}$^{1}$,
  \textbf{Zheng Zhu}$^{3}$,
  \textbf{Jie Zhou}$^{1}$,
  \textbf{Jiwen Lu}$^{1}$\\
  \\
  $^{1}$Tsinghua University \qquad
  $^{2}$BUPT \qquad
  $^{3}$GigaAI \qquad
}

\begin{document}
\maketitle


\begin{abstract}
    Spatial generalization is critical for imitation-learned manipulation policies, but achieving it typically requires scaling demonstrations across diverse object poses, robot configurations, and camera viewpoints. Data augmentation from a few source demonstrations offers a practical alternative to costly real-world collection. Simulation-based augmentation can create controllable variation, but requires complex environment and object setup and may introduce a sim-to-real gap. Recent real-to-real methods avoid these issues by jointly editing 3D observations and action trajectories from real demonstrations, yet they still rely on strong 3D scene parsing and geometry completion, and often produce observations tailored to 3D pointcloud policies rather than RGB-based 2D policies. We propose R2RDreamer, a real-to-real demonstration augmentation framework that preserves the geometric consistency of 3D action-observation editing while moving visual completion to 2D video space. Specifically, R2RDreamer first performs lightweight 3D augmentation by editing incomplete object pointclouds and end-effector trajectories in a shared 3D frame; it then projects the edited scene into masked image-space control videos with occlusion-aware reasoning and uses a dense-control image-to-video model to complete temporally coherent RGB observations. Experiments on spatially shifted manipulation tasks with both 2D diffusion-style policies and vision-language-action policies show that R2RDreamer improves spatial generalization from limited source demonstrations, with analyses validating the contributions of 3D editing, occlusion-aware projection, and video completion. Project page: \href{https://r2rdreamer.github.io/}{https://r2rdreamer.github.io/}.

\end{abstract}

\keywords{Robotic Manipulation, Data Augmentation, Video Generation} 

\section{Introduction}




Robotic manipulation with visuomotor policies has made rapid progress in recent years, from compact diffusion-style policies~\cite{chi2023diffusion,zhao2023learning} to larger vision-language-action models~\cite{brohan2023rt,kim2024openvla,black2410pi0}. However, imitation learning remains fundamentally data intensive, especially when policies must generalize spatially across diverse object poses, robot configurations, and camera viewpoints. Such \textbf{\emph{spatial generalization}} is critical for robust deployment, yet collecting demonstrations by repeatedly relocating objects and re-teleoperating the same skill is expensive and redundant: the contact-rich skill often stays the same, while most human effort is spent covering geometric variation. Data augmentation from a small number of source demonstrations therefore provides a practical route to scaling manipulation data, aiming to expand existing observation-action pairs into spatially diverse training examples while preserving the demonstrated task~\cite{lin2024data}.

\begin{figure*}[t]
    \centering
    \includegraphics[width=\textwidth]{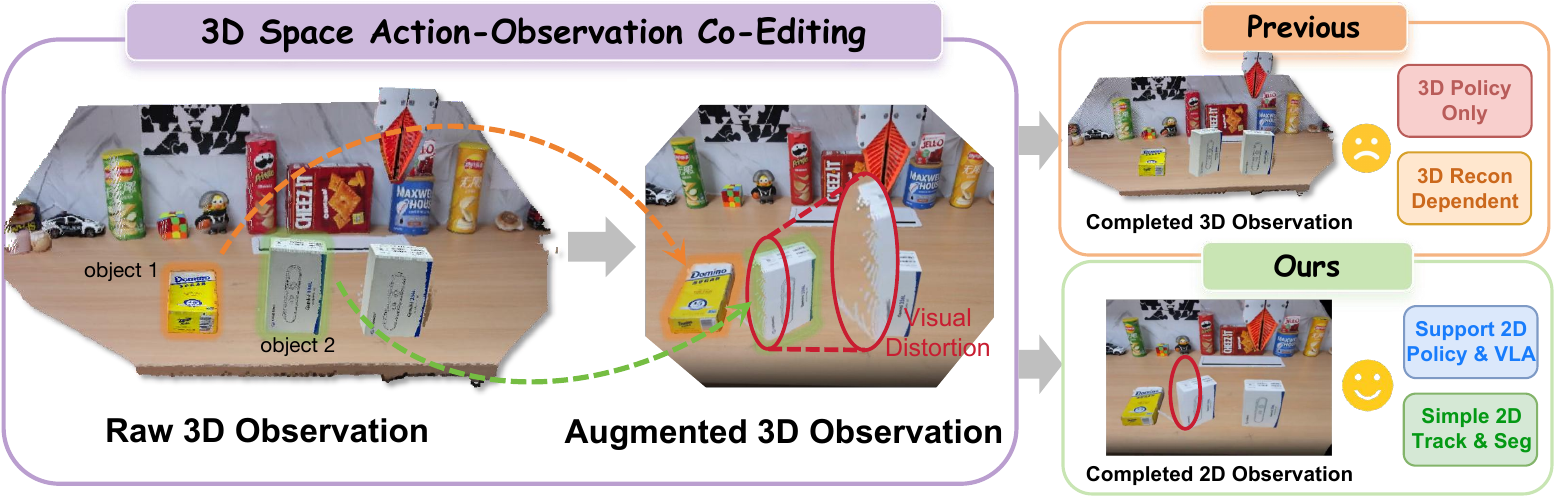}
    \caption{
    R2RDreamer keeps the geometric consistency of real-to-real 3D observation-action co-editing, but shifts visual repair from 3D geometry completion to occlusion-aware 2D control projection and video completion.
    }
    \vspace{-2mm}
    \label{fig:r2r_comparison}
\end{figure*}

Existing augmentation methods face a tradeoff between scalability and geometric fidelity. Simulation-based systems, such as MimicGen and its variants~\cite{mandlekar2023mimicgen,hoque2024intervengen,garrett2024skillmimicgen,jiang2024dexmimicgen}, can synthesize diverse executions from source demonstrations, but require simulation assets, object and environment setup, and may introduce a sim-to-real gap. Recent real-to-real methods avoid simulator construction by editing real RGB-D demonstrations directly: they parse observations into 3D, jointly transform object states and end-effector trajectories, and train policies on the augmented observation-action pairs. DemoGen~\cite{xue2025demogen} demonstrates the promise of this paradigm for pointcloud policies, while R2RGen~\cite{xu2025r2rgen} further shows that joint 3D editing can support richer spatial changes, camera setups, and multi-object interaction structures. The central challenge, however, is visual distortion after editing incomplete real observations, as illustrated in Fig.~\ref{fig:r2r_comparison}. RGB-D sensors only observe visible surfaces from the original camera trajectory; after a large spatial edit, surfaces that should appear in the new view may be missing, while surfaces that should be hidden may remain visible. Existing real-to-real methods therefore introduce additional assumptions to reduce this visual mismatch: DemoGen limits the augmentation range and task structure, whereas R2RGen relies on stronger 3D scene parsing, object geometry completion, and camera-aware 3D post-processing. These requirements depend on accurate templates, 6-DoF tracking, SAM 3D-style completion, or similar perception modules, which become restrictive for low-quality observations, fast motion, clutter, articulated objects, and deformable objects. Moreover, the resulting observations are primarily designed for 3D pointcloud policies, leaving them poorly aligned with the broad family of RGB-based 2D policies and VLAs. Therefore, scalable real-to-real augmentation that preserves 3D action-observation consistency while producing realistic RGB observations for 2D policies remains under-explored.

We propose \textbf{R2RDreamer}, a real-to-real demonstration augmentation framework that keeps spatial augmentation in 3D while formatting the final augmented data in 2D video space. R2RDreamer preserves the key geometric principle of prior real-to-real methods: object states and robot actions must be edited jointly in a shared 3D frame. Unlike prior methods, however, it does not require the edited 3D scene to be a complete policy-ready observation. Instead, R2RDreamer uses lightweight segmentation and tracking to obtain incomplete task-relevant object pointclouds, reconstructs reliable robot geometry from the robot model, and jointly edits these 3D observations with end-effector trajectories. It then projects the edited 3D scene into masked image-space control videos using occlusion-aware reasoning over self-occlusion and external occlusion, and uses a dense-control image-to-video completion model to synthesize temporally coherent RGB observations paired with the edited actions. This design shifts the difficult visual completion problem from task-specific 3D reconstruction to scalable 2D video completion, enabling augmented data for both compact 2D visuomotor policies and larger vision-language-action policies. Our experiments on spatially shifted manipulation tasks analyze the roles of 3D editing, occlusion-aware projection, and video completion, and show that R2RDreamer improves spatial generalization from limited source demonstrations while avoiding the strong 3D completion assumptions of previous real-to-real augmentation pipelines.

\section{Related Work}
    \textbf{Robotic Demonstration Augmentation:}
Imitation learning remains strongly constrained by demonstration diversity, especially under spatial shifts in object pose, robot configuration, and camera viewpoint~\cite{zhao2023learning,chi2023diffusion,lin2024data}. Simulation-based methods such as MimicGen and its extensions~\cite{mandlekar2023mimicgen,hoque2024intervengen,garrett2024skillmimicgen,jiang2024dexmimicgen} synthesize executions by replaying or recomposing demonstrations in simulated scenes, and generative-simulation pipelines use language or vision-language models to construct tasks and assets~\cite{wang2023gensim,wang2023robogen,hua2024gensim2}; both provide controllable variation but require simulation setup and may suffer sim-to-real gaps. Other systems reduce robot data requirements from human videos or rendered reconstructions~\cite{duan2023ar2,lepert2025phantom,yu2025real2render2real}, yet depend on accurate object, hand, or scene reconstruction. Real-to-real methods avoid simulator construction by editing real demonstrations directly: DemoGen~\cite{xue2025demogen} augments pointcloud observations and actions, and R2RGen~\cite{xu2025r2rgen} supports richer 3D spatial transformations and interaction structures. These methods still rely on strong 3D perception or geometry completion to reduce visual mismatch after editing incomplete observations, and their outputs are mainly suited to 3D pointcloud policies. R2RDreamer keeps joint 3D observation-action editing but shifts visual completion to 2D video space, enabling augmented data for RGB-based 2D policies and VLAs.

\textbf{Video Models for Robotics:}
Video generation models can synthesize temporally coherent RGB observations under visual, geometric, or action-related conditions. Recent methods use them for offline robot-data augmentation through appearance, viewpoint, embodiment, or modality transfer~\cite{liu2025robotransfer,tong2025fidelity,qian2025wristworld,li2025mimicdreamer}, or build larger embodied data engines with 3D reconstruction, view transfer, and motion planning~\cite{zhao2025real2edit2real,team2025gigaworld}. Video models are also studied as interactive world models for planning, policy decoding, evaluation, or virtual rollout~\cite{yang2023unisim,du2023unipi,zhou2024robodreamer,jang2025dreamgen,liao2025genie,jiang2025enerverse,xiao2025worldenv,kim2026cosmospolicy,chen2026abotphysworld}. These works show the promise of scalable embodied priors, but often generate full interaction dynamics, rely on dense reconstruction or action-conditioned simulation, or focus on appearance and viewpoint transfer. R2RDreamer uses video models in a more targeted role: explicit 3D editing provides consistent action labels and projected controls, while dense-control video completion repairs the masked visual consequences of real-to-real spatial augmentation.

\section{Approach}
    \subsection{Problem Statement}\label{sec:problem}
\textbf{Visuomotor policy learning:}
We consider imitation learning for a visuomotor policy $\pi$, which maps visual observations and optional language instructions to robot actions.
At time $t$, the robot observes $o_t=(I_t,D_t)$ and executes action $a_t=(\mathbf{A}_t^{ee},a_t^{grip})$, where $\mathbf{A}_t^{ee}\in \mathrm{SE}(3)$ is the end-effector pose and $a_t^{grip}$ is the gripper command.
Depth $D_t$ can be lifted to a pointcloud $P_t$ for geometry processing, but the target policy may consume only RGB frames or video histories.
Each task can optionally carry a language instruction $l$.
Compact visuomotor policies use $\pi(o_t)$, while VLA-style policies use $\pi(o_t,l)$.
Since collecting diverse demonstrations to cover different spatial distribution of objects is very time-consuming, our work aims to augment limited human demonstrations into spatially diverse ones.
Given a source demonstration
\begin{equation}
    D_s=\left(l,\{(I_t,D_t,a_t)\}_{t=1}^{H}\right),
\end{equation}
R2RDreamer produces augmented demonstrations
\begin{equation}
    \hat{D}^j=\left(l,\{(\hat{I}_t^j,\hat{a}_t^j)\}_{t=1}^{\hat{H}_j}\right),
\end{equation}
which preserve the task semantics and contact structure of $D_s$ while varying object poses, robot configurations, and camera viewpoints.
The final data are RGB/video observations paired with edited actions, so they can train both compact 2D visuomotor policies and larger VLA-style policies.

\textbf{Why 3D is still needed:}
Image-only editing cannot guarantee that the action reaches the edited object state.
Following the real-to-real principle of R2RGen~\cite{xu2025r2rgen}, R2RDreamer therefore edits task-relevant object observations and actions in a shared 3D frame.
However, it releases the strong assumption that the edited 3D scene must be complete enough for direct policy input.
Incomplete 3D observations are used as control geometry; unknown or invalid pixels are left black in the projected control video and repaired by the video model.

\begin{figure*}[t]
    \centering
    \includegraphics[width=\textwidth]{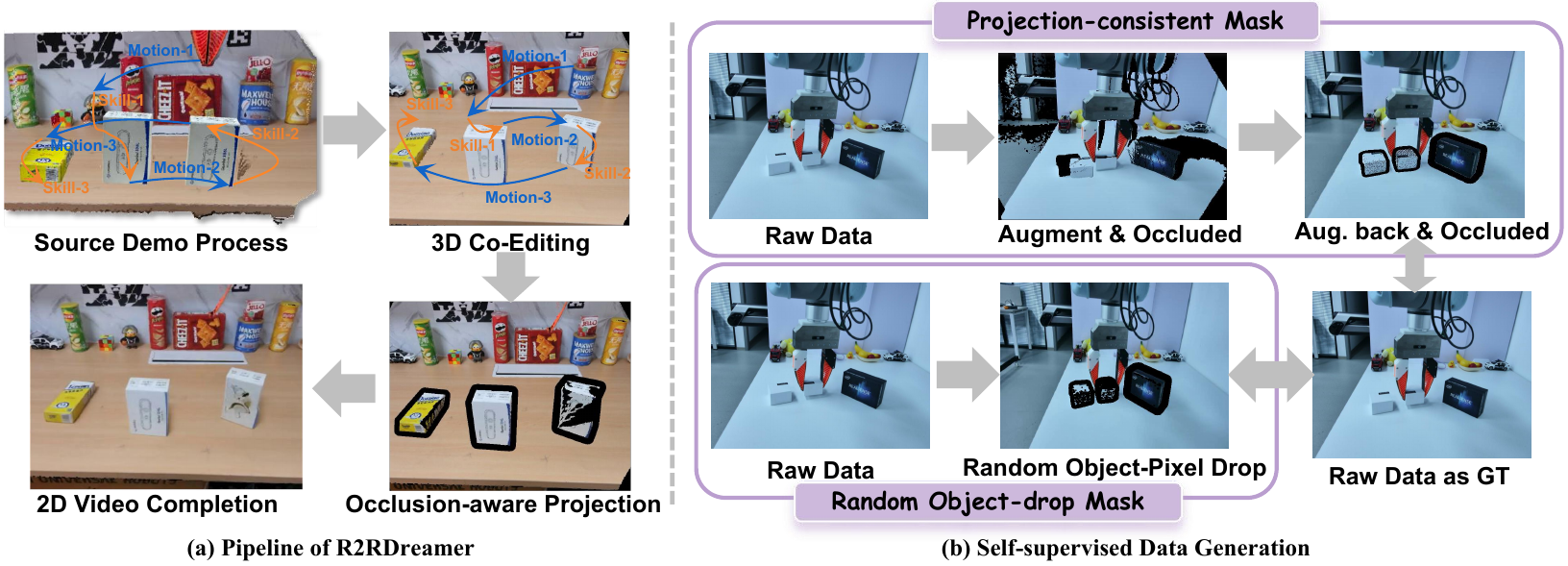}
    \caption{
    R2RDreamer pipeline.
    (a) Demonstration augmentation: source RGB-D demonstrations are segmented and tracked, co-edited with actions in 3D, projected into masked 2D control videos, and completed into RGB observations for policy training.
    (b) Self-supervised training pairs for video completion: the top branch builds projection-consistent masks through augment--occlude--augment-back using measured or estimated depth, while the bottom branch builds random object-drop masks from tracked object pixels.
    }
    \label{fig:projection_pipeline}
\end{figure*}

\subsection{3D Co-Editing with Occlusion-Aware Projection}\label{sec:3d_projection}
\textbf{Lightweight 3D co-editing:}
R2RDreamer uses a simplified real-to-real editing stage, shown in Fig.~\ref{fig:projection_pipeline}(a).
Given the source RGB-D video, task-relevant objects are segmented in the first frame and tracked with SAM2~\cite{ravi2024sam}.
Let $t$ index the frame and $k$ index the task-relevant object.
Each mask $M_t^k$ selects an incomplete object pointcloud $P_t^k$ from the depth-lifted observation.
In prior real-to-real methods, such partial pointclouds must become complete policy observations because the edited pointcloud is directly fed to the policy; otherwise missing newly visible surfaces or invalid originally visible surfaces become training artifacts.
R2RDreamer releases this assumption and does not replace partial observations with completed meshes, scanned templates, or SAM-3D-style geometry.
The robot arm is reconstructed from the robot model and kinematic state, since its geometry is known and should remain crisp in the projected control.

All editable elements are represented in a shared world frame.
We partition the source trajectory into skill segments, where the gripper executes object interaction, and motion segments, which connect adjacent skills.
For a skill interacting with object group $G_i$, R2RDreamer samples a transform $\mathbf{T}_i$ and applies it to the relevant object pointclouds and end-effector poses:
\begin{equation}
    \hat{P}_t^k=\mathbf{T}_i P_t^k,\quad
    \hat{\mathbf{A}}_t^{ee}=\mathbf{T}_i\mathbf{A}_t^{ee},\quad
    k\in G_i .
\end{equation}
This preserves the relative observation-action geometry of the contact-rich skill while moving it to new spatial locations.
The gripper command is preserved; motion segments are generated by interpolation or motion planning, and in-hand objects follow the relative end-effector transform.
Thus motion expands spatial distribution while the demonstrated interaction remains executable.
Because the pointcloud is only an intermediate carrier for co-editing and a projected condition for 2D completion, incomplete geometry is acceptable if unreliable pixels are masked.
This reduces reliance on strong 3D reconstruction models like FoundationPose~\cite{wen2024foundationpose} and SAM 3D~\cite{chen2025sam}, and makes rigid, articulated, and deformable objects follow the same partial-observation interface; previous methods struggle on non-rigid objects because reliable completed geometry is difficult.

\textbf{Occlusion-aware projection:}
After co-editing, the incomplete pointcloud is projected into the target camera as a 2D control video, with visually unreliable regions left black for video completion.
Let $\Pi$ denote camera projection with z-buffer rendering, and let $\mathcal{S}(\cdot)$ return the occupied image support of a projected point set.
We first render an unmasked control frame from the edited objects, model-based robot pointcloud, and unchanged environment pointcloud:
\begin{equation}
    \tilde{\mathbf{C}}_t = \Pi\!\left(\hat{P}_t^1,\ldots,\hat{P}_t^K,\hat{P}_t^{robot},P_t^{env}\right),
\end{equation}
where pixels not hit by any projected point are initialized as black.
Thus $\tilde{\mathbf{C}}_t$ is already a sparse projected control frame; self- and external-occlusion further black out projected pixels that are geometrically unsupported.

Self-occlusion removes source-visible 3D points that should become hidden after object motion.
For object $k$, $\texttt{HPR}$ keeps points visible from the target camera but cannot model the unobserved backside of sparse RGB-D pointclouds.
We place a thin backside proxy $B_\delta(\hat{P}_t^k)$ behind the originally visible surface before transformation; after a large rotation, it occludes source-visible samples that should now be hidden.
Because a single-pixel z-buffer is brittle under pointcloud holes and small projection errors, we test occlusion over a local image patch:
\begin{equation}
    \texttt{CamVis}(\hat{P}_t^k)
    =
    \left\{
    p\in\texttt{HPR}(\hat{P}_t^k)
    \;\middle|\;
    z(p) \le
    \min_{b\in B_\delta(\hat{P}_t^k),\ \|\Pi(b)-\Pi(p)\|_\infty\le r}
    z(b)
    \right\},
    \label{eq:camvis}
\end{equation}
where $z(\cdot)$ is camera depth and $r$ is the local z-buffer radius.
The self-occlusion holes are the projected object footprint not supported by trusted camera-visible points:
\begin{equation}
    \mathbf{Q}_t^{self,k}
    =
    \mathcal{S}\!\left(\Pi(\hat{P}_t^k)\right)
    \setminus
    \mathcal{S}\!\left(\Pi(\texttt{CamVis}(\hat{P}_t^k))\right).
    \label{eq:occlusion_operator}
\end{equation}

External occlusion covers uncertainty from contacts, background regions, and partial occluders not fully reconstructed in 3D.
We approximate each projected object boundary with a polygon, dilate it by object size, and use the outside ring as $\mathbf{Q}_t^{ext}$.
The final black-filled projected control frame $\mathbf{C}_t$ is
\begin{equation}
    \mathbf{Q}_t = \left(\bigcup_k \mathbf{Q}_t^{self,k}\right) \cup \mathbf{Q}_t^{ext},
    \qquad
    \mathbf{C}_t=(1-\mathbf{Q}_t)\odot \tilde{\mathbf{C}}_t + \mathbf{Q}_t\odot \mathbf{0}.
    \label{eq:black_fill}
\end{equation}
Stacking $\{\mathbf{C}_t\}_{t=1}^{H}$ gives the masked control video; $\mathbf{Q}_t$ is reused below for self-supervised training.

\subsection{Projection-Controlled Video Completion Model}\label{sec:video_completion}
\begin{figure*}[t]
    \centering
    \includegraphics[width=\linewidth]{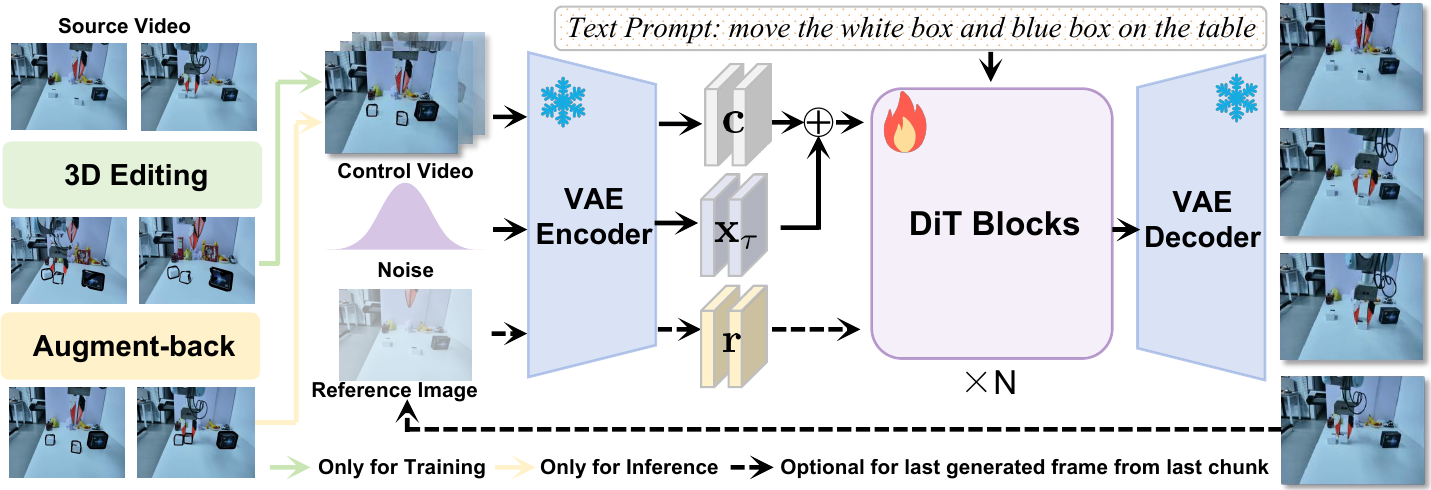}
    \caption{Projection-controlled video completion model and training \& inference pipeline.}
    \label{fig:video_completion}
\end{figure*}

\textbf{Control-augmented IT2V model:}
As illustrated in Fig.~\ref{fig:video_completion}, R2RDreamer builds on WAN2.2 IT2V~\cite{team2025wan}, which generates future video from text and an optional reference image.
We add a dense control branch for $\mathbf{C}_{1:H}$, which is temporally and pixel-wise aligned with the target video.
The clean and control videos are encoded by the same WAN VAE as latents $\mathbf{x}_1$ and $\mathbf{c}$; after sampling $\mathbf{x}_0\sim\mathcal{N}(0,\mathbf{I})$ and $\tau\sim\mathcal{U}(0,1)$, we form $\mathbf{x}_\tau=(1-\tau)\mathbf{x}_0+\tau\mathbf{x}_1$.
Each token of $\mathbf{x}_\tau$ and $\mathbf{c}$ is channel-concatenated, linearly projected, and fed to the WAN DiT with text and the reference-image latent condition.
The flow-matching loss is
\begin{equation}
    \mathcal{L}_{fm}
    =
    \mathrm{E}_{\mathbf{x}_0,\mathbf{x}_1,\tau}
    \left[
    \left\|
    \mathbf{w}(\mathbf{q})\odot
    \left(v_\theta([\mathbf{x}_\tau;\mathbf{c}],\tau,l,\mathbf{r})-(\mathbf{x}_1-\mathbf{x}_0)\right)
    \right\|_2^2
    \right],
    \label{eq:flow_matching}
\end{equation}
where $l$ is the language instruction, $\mathbf{r}$ is the optional reference-image latent, $\mathbf{q}$ is $\mathbf{Q}_{1:H}$ at VAE resolution, and $\mathbf{w}(\mathbf{q})=1+\lambda(1-\mathbf{q})$ gives non-black regions a larger penalty to discourage modifying pixels already specified by the control video.
At inference, long videos are generated chunk by chunk.
The first chunk is conditioned on language and projected video control only, while each subsequent chunk additionally uses the last generated frame of the previous chunk as the reference image to improve temporal continuity.
When strict preservation of projected pixels is desired, we optionally compose the final frame by taking non-black pixels from the control video and generated pixels from the masked regions, with boundary blending near the mask edge.

\textbf{Self-supervised training data:}
Training the completion model requires pairs of masked control videos and clean target videos, but real 3D co-editing does not provide ground-truth completed RGB observations after spatial augmentation.
We therefore construct mask pairs self-supervisedly from ordinary RGB videos: the original video serves as the clean target, and synthetic masks simulate the regions that would become unreliable after real-to-real editing.
For a clean video $V=\{I_t\}_{t=1}^{H}$, we track task-relevant objects to build temporally aligned masks.

The first pair type is the projection-consistent mask, shown in Fig.~\ref{fig:projection_pipeline}(b, top).
This construction requires object pointclouds, so depth is optional and can be estimated by a monocular depth estimator~\cite{chen2025video} when not available.
For each object track in a clip, we lift the masks using measured or estimated depth, sample one transform $\mathbf{T}^k$, and reuse it across all frames, then apply frame-wise self-occlusion under the current camera pose and map surviving points back.
This video-level augment--occlude--augment-back procedure preserves the original video as the target while producing temporally coherent projection-consistent holes; external masks are added by polygon dilation.
For object $k$, this corruption is summarized as
\begin{equation}
    \bar{\mathbf{Q}}_t^{self,k}
    =
    \mathcal{S}\!\left(\Pi(P_t^k)\right)
    \setminus
    \mathcal{S}\!\left(\Pi((\mathbf{T}^k)^{-1}\texttt{CamVis}(\mathbf{T}^kP_t^k))\right),
\end{equation}
where $\mathbf{T}^k$ is shared across the clip for object $k$; compared with Eq.~\ref{eq:occlusion_operator}, the surviving points are mapped back before forming holes in the original image.

Projection-consistent masks cover the dominant geometric artifacts, while real co-editing can also produce irregular holes from imperfect segmentation and approximate occlusion hyperparameters.
The second pair type is therefore the random object-drop mask, shown in Fig.~\ref{fig:projection_pipeline}(b, bottom), which sets a sampled percentage of tracked object pixels as holes.
For either pair type, we reuse the black-filling rule in Eq.~\ref{eq:black_fill}: $\bar{\mathbf{C}}_t=(1-\bar{\mathbf{Q}}_t)\odot I_t+\bar{\mathbf{Q}}_t\odot\mathbf{0}$.
For projection-consistent pairs, $\bar{\mathbf{Q}}_t=(\cup_k\bar{\mathbf{Q}}_t^{self,k})\cup\bar{\mathbf{Q}}_t^{ext}$; for object-drop pairs, $\bar{\mathbf{Q}}_t$ is the sampled drop mask.
This training requires no actions or completed 3D geometry.
After training, the model takes the masked control video produced by 3D co-editing and outputs completed RGB observations, which are paired with the edited actions as the final augmented demonstrations.

\section{Experiment}
    \subsection{Experimental Setup}\label{sec:exp_setup}
\textbf{Tasks and hardware.}
We evaluate R2RDreamer on four real-world manipulation tasks that require spatial generalization from limited demonstrations.
The suite includes rigid-object and non-rigid/articulated-object manipulation, since the latter makes methods depending on complete 3D object geometry restrictive.
Each task is evaluated under held-out object locations that differ from the source demonstrations.
Experiments use a real robot platform with an RGB-D camera and a parallel-jaw gripper; R2RDreamer uses RGB-D observations for augmentation, while final policies consume RGB observations.
Appendix~\ref{app:tasks} provides task descriptions.

\textbf{Policies.}
We test Diffusion Policy (DP)~\cite{chi2023diffusion}, a compact visuomotor policy using RGB observations and proprioception, and $\pi_0$~\cite{black2410pi0}, a vision-language-action policy that also conditions on the instruction $l$.
This comparison tests whether the same augmented demonstrations benefit both conventional 2D visuomotor policies and larger VLA-style models.

\textbf{Baselines and metrics.}
For each policy, we evaluate one-source imitation, one-source imitation augmented by R2RDreamer, and human-demonstration baselines with 5, 15, and 30 source demonstrations.
This protocol measures how much spatial diversity R2RDreamer adds to a fixed source demonstration budget relative to collecting additional real demonstrations.
The main metric is real-world success rate over held-out trials, and qualitative results visualize the edited 3D observation, masked projected control video, and completed RGB video.
Video completion training details are provided in Appendix~\ref{app:video_training_data}.

\subsection{Main Results}\label{sec:main_results}

\begin{table*}[t]
    \centering
    \caption{
    Real-world spatial-generalization results for 2D policies.
    Success rate is reported on held-out task configurations.
    }
    \label{tab:main_results}
    \small
    \setlength{\tabcolsep}{4pt}
    \newcolumntype{Y}{>{\centering\arraybackslash}p{0.075\textwidth}}
    \begin{tabular*}{\textwidth}{@{\extracolsep{\fill}}l|YY|YY|YY|YY}
        \toprule
        \multicolumn{1}{c|}{\multirow{2}{*}{Training data}}
        & \multicolumn{2}{c|}{Pot-Food}
        & \multicolumn{2}{c|}{Hang-Cup}
        & \multicolumn{2}{c|}{Build-Bridge}
        & \multicolumn{2}{c}{Cover-Object} \\
        & DP & $\pi_0$ & DP & $\pi_0$ & DP & $\pi_0$ & DP & $\pi_0$ \\
        \midrule
        $1$ Source
        & 3.13 & 3.13 & 3.13 & 3.13 & 3.13 & 3.13 & 3.13 & 3.13 \\
        + \textbf{R2RDreamer}
        & \textbf{21.9} & \textbf{21.9} & \textbf{28.1} & \textbf{28.1} & \textbf{21.9} & \textbf{25.0} & \textbf{40.6} & \textbf{46.9} \\
        \midrule
        \textcolor{gray}{$5$ Source}
        & \textcolor{gray}{6.25} & \textcolor{gray}{6.25} & \textcolor{gray}{9.38} & \textcolor{gray}{9.38} & \textcolor{gray}{6.25} & \textcolor{gray}{6.25} & \textcolor{gray}{9.38} & \textcolor{gray}{12.5} \\
        \textcolor{gray}{$15$ Source}
        & \textcolor{gray}{18.8} & \textcolor{gray}{25.0} & \textcolor{gray}{25.0} & \textcolor{gray}{31.3} & \textcolor{gray}{15.6} & \textcolor{gray}{21.9} & \textcolor{gray}{31.3} & \textcolor{gray}{40.6} \\
        \textcolor{gray}{$30$ Source}
        & \textcolor{gray}{43.8} & \textcolor{gray}{50.0} & \textcolor{gray}{46.9} & \textcolor{gray}{59.4} & \textcolor{gray}{31.3} & \textcolor{gray}{37.5} & \textcolor{gray}{50.0} & \textcolor{gray}{62.5} \\
        \bottomrule
    \end{tabular*}
\end{table*}

\begin{figure*}[t]
    \centering
    \includegraphics[width=\linewidth]{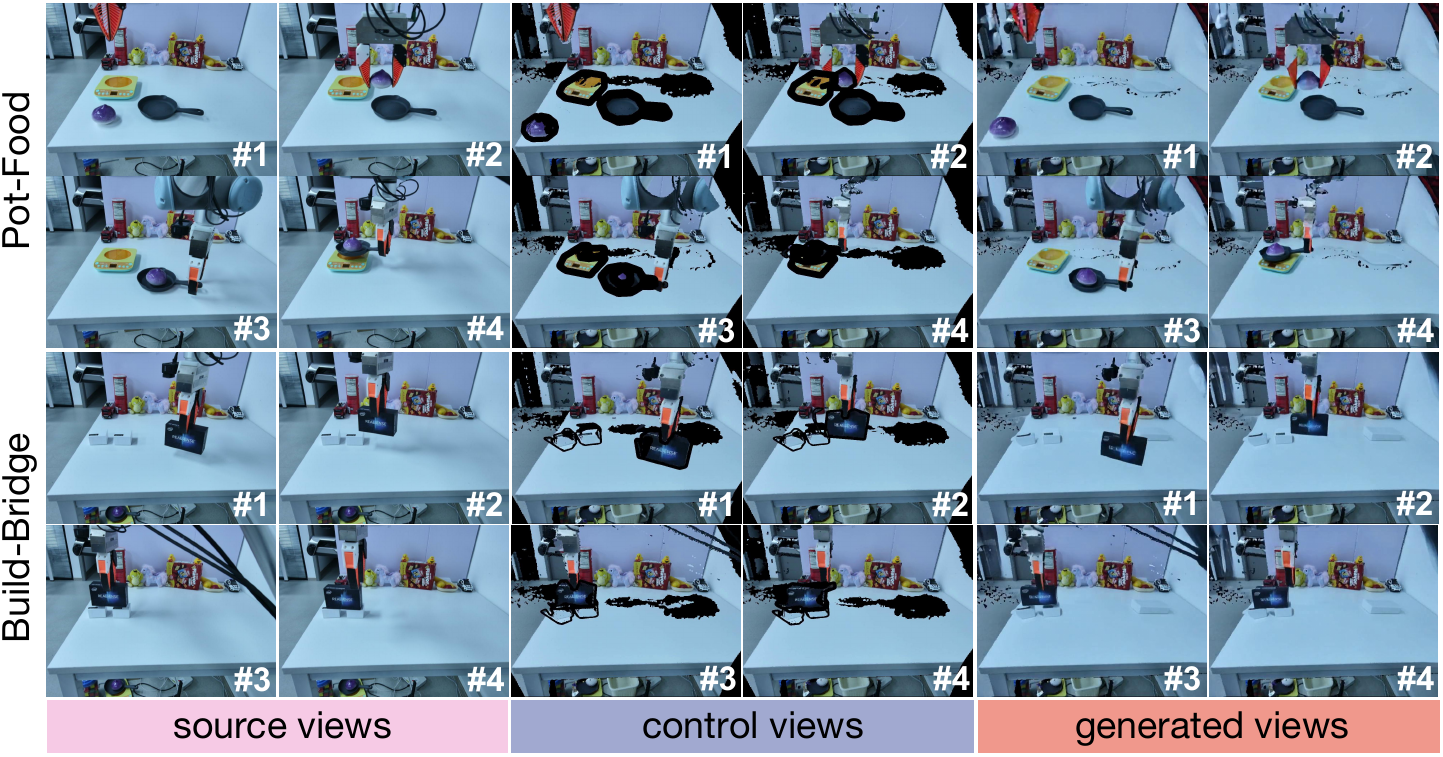}
    \caption{Qualitative visualization of 3D co-editing sources views, masked projection results as control views, and video completion as generated views.}
    \label{fig:main_qual}
\end{figure*}

Table~\ref{tab:main_results} compares R2RDreamer with policies trained on different numbers of human demonstrations.
With one source demonstration, both DP and $\pi_0$ largely overfit to the demonstrated configuration and achieve only $3.13\%$ success on held-out spatial layouts.
Augmenting the same source demonstration with R2RDreamer consistently improves performance across all four tasks and both policy families, reaching $21.9$--$40.6\%$ for DP and $21.9$--$46.9\%$ for $\pi_0$.
These gains are well above the 5-source human-data baseline and are comparable to or better than the 15-source baseline in most settings, although the 30-source baseline remains stronger.
The result indicates that R2RDreamer adds substantial spatial diversity to a fixed source demonstration budget, while still leaving room for complementary gains from collecting more real demonstrations.
Because the final augmented data are completed RGB observations paired with edited actions, the same pipeline benefits both DP without language and $\pi_0$ with language.

Figure~\ref{fig:main_qual} illustrates the main augmentation path.
The edited 3D observations preserve action-relevant contact geometry, the projected controls leave unsupported pixels black, and the completed videos recover temporally coherent RGB observations around edited objects, robot links, and contact regions.
This visual sequence supports the interpretation that policy gains come from action-consistent spatial diversity rather than appearance-only perturbations.

\subsection{Ablation Study}\label{sec:ablation}

\begin{figure*}[t]
    \centering
    \includegraphics[width=\linewidth]{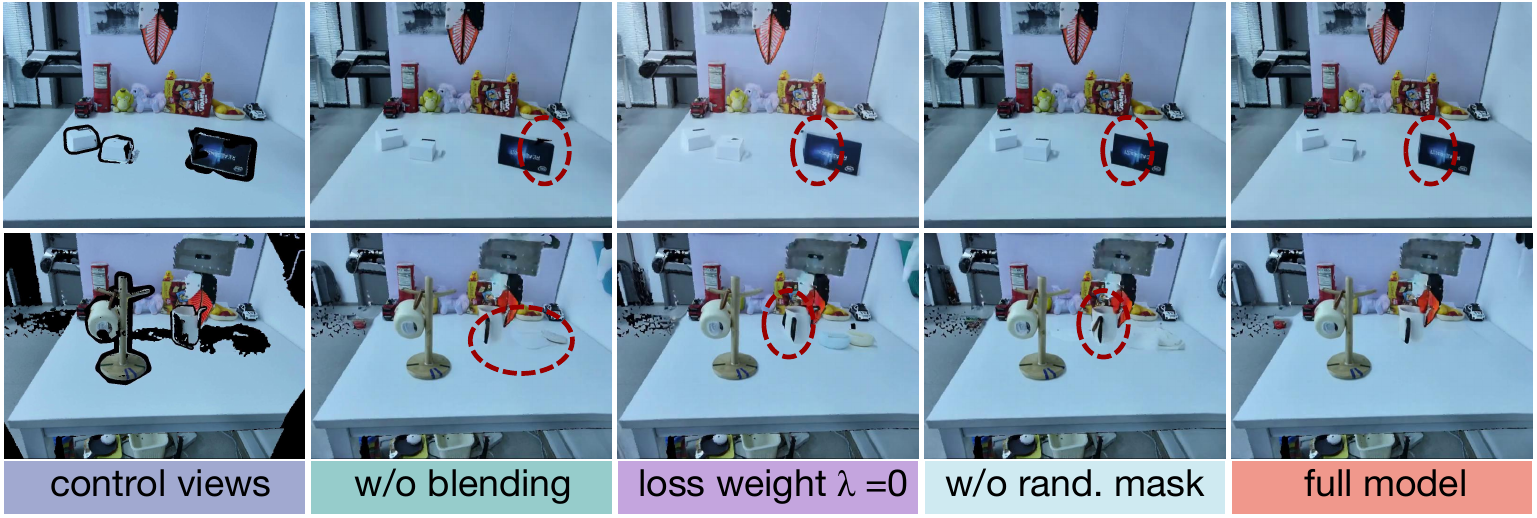}
    \caption{Qualitative ablations of training masks, inference blending, and loss weight $\lambda$.}
    \label{fig:ablation_qual}
\end{figure*}

\textbf{Mask-pair construction.}
Removing random object-drop pairs reduces robustness to irregular missing regions caused by segmentation errors, sparse depth, and imperfect occlusion hyperparameters.
The model then tends to overfit to clean geometric masks and can leave artifacts when black regions are fragmented.
Removing projection-consistent pairs has a different effect: the model sees random local corruption but loses exposure to temporally coherent holes caused by 3D motion and self-occlusion, weakening completion near object boundaries and newly revealed surfaces.

\textbf{Inference composition and loss weighting.}
Inference-time blending helps preserve reliable non-black control pixels.
Without blending, the video model may slightly alter pixels that should stay fixed, introducing small geometry drift around the robot arm, object contact point and background artifacts.
However, excessive hard composition can create visible seams near mask boundaries, so boundary blending is used when projected and generated appearances differ.
The loss weight $\lambda$ controls the same tradeoff during training: small values allow the model to modify specified control regions too freely, while overly large values can preserve projected artifacts and reduce harmonization inside black regions.
We choose $\lambda$ by visually balancing control preservation in non-black regions and realism inside black regions.

\section{Limitation}
    R2RDreamer avoids complete 3D object reconstruction, but it still depends on clean RGB-D geometry and reliable 2D segmentation/tracking; noisy depth, pointcloud artifacts, or failed SAM2 masks can degrade the projected control video and final completion.
Its control branch is also trained on far less data than the WAN foundation model, so stronger out-of-distribution completion will likely require broader manipulation videos and more diverse masked-video patterns; otherwise, completion can still distort geometry for object shapes far outside the training distribution.

\section{Concluding Remark}
    We presented R2RDreamer, a real-to-real demonstration augmentation framework for spatially generalized 2D manipulation policies.
It keeps the essential 3D consistency of observation-action editing, but treats incomplete edited pointclouds as projected controls and repairs unreliable regions with dense-control video completion.
This reduces reliance on complete 3D object geometry and produces augmented demonstrations usable by both compact visuomotor policies and VLA-style policies.
Future work can further improve long-horizon temporal consistency and study how the completed videos scale with larger multi-task policy training.


\bibliography{main}  

\newpage
\appendix
\section*{Appendix}
\section{Task Suite}
\label{app:tasks}

\begin{figure}[H]
    \centering
    \includegraphics[width=\textwidth]{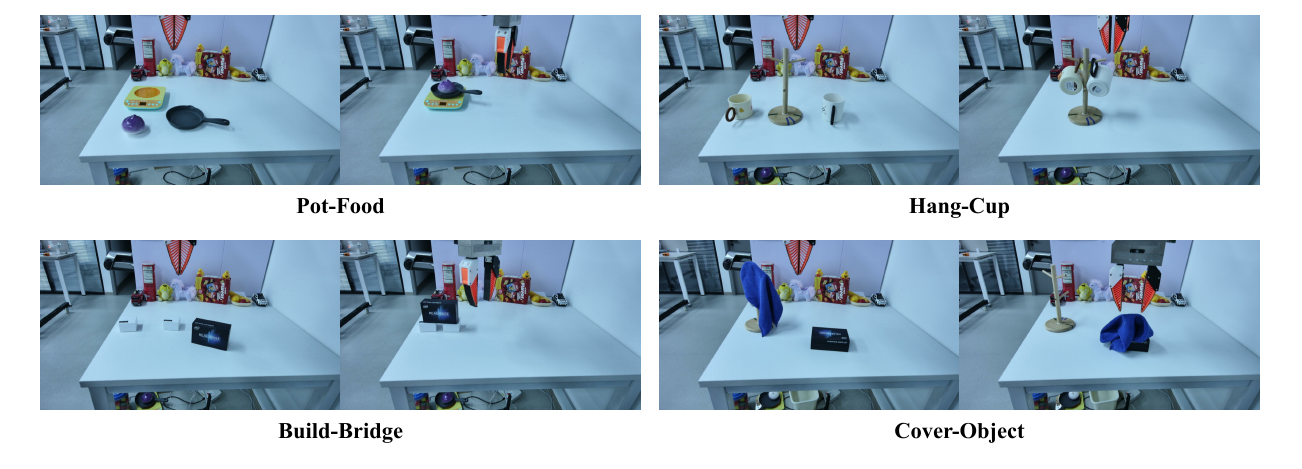}
    \caption{
    Visualization of the four real-world manipulation tasks used for spatial-generalization evaluation.
    Each task is shown with a start frame and an end frame from a successful rollout, illustrating the initial object arrangement and the final task relation that the augmented RGB-action data must preserve.
    }
    \label{fig:app_tasks}
\end{figure}

This appendix provides additional descriptions of the four real-world manipulation tasks used in Sec.~\ref{sec:exp_setup}.
The tasks are selected to evaluate spatial generalization from limited source demonstrations across multi-object rigid manipulation and non-rigid object interaction.
Across tasks, evaluation configurations vary the spatial arrangement of task-relevant objects relative to the source demonstrations.
Fig.~\ref{fig:app_tasks} shows the start and end frames for each task.
These task descriptions complement the main paper's quantitative results by clarifying the object relations and interaction outcomes that the augmented RGB-action data must preserve.

\paragraph{Pot-Food.}
The task contains an onion, a pot, and a stove.
The robot first grasps the onion and places it into the pot.
It then grasps the pot and moves it onto the stove.
Success is achieved when the onion is inside the pot and the pot is placed on the stove at the end of the rollout.
This task tests whether augmentation preserves a temporally ordered multi-object routine: the onion must be placed into the pot before the pot is moved to its final target location.

\paragraph{Hang-Cup.}
The task contains two cups and a rack with multiple hanging positions.
The robot picks up the first cup and hangs it on the rack, then repeats the operation for the second cup while placing it at a different valid rack position.
Success is achieved when both cups are stably hung without occupying the same target location.
This task stresses spatial generalization over repeated pick-and-place interactions with similar objects, where object identity and final relative placement both matter.

\paragraph{Build-Bridge.}
The task contains two white boxes and one black box.
The robot first places the two white boxes at a proper distance, then grasps the black box and places it across the white boxes to form a bridge.
Success is achieved when the black box is supported by both white boxes in a stable bridge structure.
This task requires preserving a higher-order spatial constraint: independently plausible object placements are not sufficient unless the spacing between the white boxes and the pose of the black box jointly satisfy the bridge geometry.

\paragraph{Cover-Object.}
The task contains a blue cleaning cloth hanging on a rack and a black box on the table.
The robot removes the cloth from the rack and places it over the black box.
Success is achieved when the cloth covers the black box after release.
This task evaluates R2RDreamer in a setting where complete 3D object geometry is especially difficult to obtain, since the cloth can deform and self-occlude during manipulation.
It therefore highlights the motivation for using incomplete 3D editing as control and relying on 2D video completion to synthesize realistic RGB observations.

\section{Video Completion Training Data}
\label{app:video_training_data}

We train the projection-controlled video completion model in three stages.
The first stage uses ScanNet~\cite{dai2017scannet} scene data, where depth and object annotations provide clean geometry for projection-consistent mask construction.
The second stage uses BridgeData V2~\cite{walke2023bridgedata} manipulation videos; because these videos do not provide the same 3D annotations, we estimate depth with a monocular depth model and obtain object tracks with SAM2.
Finally, we perform a small amount of finetuning on our own offline robot data so that the completion model better matches the camera viewpoint, robot appearance, and object distribution used in the real-world evaluation.
Across these stages, the training pairs follow the mask constructions described in Sec.~\ref{sec:video_completion}: projection-consistent pairs provide temporally coherent geometric holes, while random object-drop pairs improve robustness to irregular missing regions.

\end{document}